\documentclass{article} % For LaTeX2e
\usepackage[table,xcdraw]{xcolor}
\usepackage{main,times}

\usepackage{amsmath,amsfonts,bm}

\def\eqref#1{equation~\ref{#1}}
\def\1{\bm{1}}

\DeclareMathAlphabet{\mathsfit}{\encodingdefault}{\sfdefault}{m}{sl}
\SetMathAlphabet{\mathsfit}{bold}{\encodingdefault}{\sfdefault}{bx}{n}

\usepackage{hyperref}
\usepackage{url}
\usepackage{subcaption}
\usepackage{graphicx}
\usepackage{booktabs}
\usepackage{multirow}
\usepackage{makecell} 

\title{CycleHOI: Improving Human-Object Interaction Detection  with Cycle Consistency of Detection and Generation}

\author{Yisen Wang, Yao Teng \& Limin Wang \\% \thanks{Corresponding author (lmwang@nju.edu.cn)} \\
State Key Laboratory for Novel Software Technology, Nanjing University, China \\
\texttt{yswang@smail.nju.edu.cn}%, tengyao19980325@gmail.com}\\
}
\iclrfinalcopy % Uncomment for camera-ready version, but NOT for submission.
\begin{document}

\maketitle

\newcommand{\hicosotatable}{
\begin{table}[ht]
% \vspace{-2mm}
\centering
\small
\renewcommand{\arraystretch}{1.2}
\setlength{\tabcolsep}{2.2pt}
\begin{tabular}{llccccccc}
\toprule
\multirow{2}{*}{Method} & \multirow{2}{*}{Backbone} & \multicolumn{3}{c}{Default} && \multicolumn{3}{c}{Known Object} \\
\cmidrule{3-5} \cmidrule{7-9}
&& Full & Rare & Non-Rare && Full & Rare & Non-Rare  \\
\midrule
HOTR \citep{hotr} & R-50 &25.10 & 17.34 & 27.42 && - & - & - \\
HOI-Trans \citep{e2e_hoi}   & R-101 & 26.61 & 19.15 & 28.84 && 29.13 & 20.98 & 31.57 \\
AS-Net \citep{setprediction_hoi} & R-50 & 28.87 & 24.25 & 30.25 && 31.74 & 27.07 & 33.14 \\
QPIC \citep{qpic_hoi}  & R-50 & \underline{29.07} & 21.85 & 31.23 && \underline{31.68} & 24.14 & 33.93 \\
SCG \citep{zhang2021spatially} & R-50 & 29.26 & 24.61 & 30.65 && 32.87 & 27.89 & 34.35 \\
MSTR \citep{kim2022mstr} & R-50 & 31.17 & 25.31 & 32.92 && 34.02 & 28.83 & 35.57 \\
SSRT \citep{iftekhar2022look}  & R-101 & 31.34 & 24.31 & 33.32 && - & - & - \\
CDN \citep{mining_hoi}  & R-101 & 32.07 & 27.19 & 33.53 && 34.79 & 29.48 & 36.38 \\
STIP \citep{zhang2022exploring} & R-50 & 32.22 & 28.15 & 33.43 && 35.29 & 31.43 & 36.45 \\
DOQ \citep{doq} & R-50 & 33.28 & 29.19 & 34.50 && - & - & - \\
UPT \citep{zhang2022efficient} & R-101 & 32.62 & 28.62 & 33.81 && 36.08 & 31.41 & 37.47 \\
DEFR \citep{jin2022overlooked} & ViT-B/16 & 32.35 & 33.45 & 32.02 && - & - & - \\
IF \citep{liu2022interactiveness} & R-50 & 33.51 & 30.30 & 34.46 && 36.28 & 33.16 & 37.21\\
GEN-VLKT \citep{GEN_VLKT} & R-101 & \underline{34.95} & 31.18 & 36.08 && \underline{38.22} & 34.36 & 39.37 \\
QAHOI \citep{chen2021qahoi} & Swin-L & 35.78 & 29.80 & 37.56 && 37.59 & 31.66 & 39.36 \\
FGAHOI \citep{ma2023fgahoi} & Swin-L & 37.18 & 30.71 & 39.11 && 38.93 & 31.93 & 41.02 \\
ViPLO \citep{ViPLO} & ViT-B/16 & 37.22 & 35.45 & 37.75 && 40.61 & 38.82 & 41.15 \\
PViC \citep{pvic} & Swin-L & \underline{44.32} & 44.61 & 44.24 && \underline{47.81} & 48.38 & 47.64 \\
\midrule
Ours (QPIC) & R-50 & 32.23{\color{blue}$\uparrow_{3.16}$} & 25.27 & 34.01 && 34.80{\color{blue}$\uparrow_{3.12}$} & 27.58 & 36.83\\
Ours (GEN-VLKT) & R-101 & 37.79{\color{blue}$\uparrow_{2.84}$} & 34.22 & 38.61 && 41.13{\color{blue}$\uparrow_{2.91}$} & 37.43 & 42.06 \\
Ours (PViC) & Swin-L & \textbf{45.71}{\color{blue}$\uparrow_{1.39}$} & \textbf{46.14} & \textbf{45.52} && \textbf{49.23}{\color{blue}$\uparrow_{1.42}$} & \textbf{49.87} & \textbf{48.96} \\
\bottomrule
\end{tabular}
\caption{
Performance of various HOI detectors on HICO-DET. We experiment on some excellent work, \underline{underline} indicate the results to be compared.} 
\label{tab:sota_hico}
% \vspace{-10mm}
\end{table}
}

\newcommand{\cocosotatable}{
\begin{table}[ht]
% \vspace{-2mm}
\begin{subtable}{.52\linewidth}
\centering
\small
\renewcommand{\arraystretch}{1.2}
\setlength{\tabcolsep}{0.4pt}
\begin{tabular}{llcc}
\toprule
Method & Backbone & $AP^{S1}_{role}$ & $AP^{S2}_{role}$ \\
\midrule
HOTR \citep{hotr} & R-50 & 55.2 & 64.4 \\
HOI-Trans \citep{e2e_hoi}   & R-101 & 52.9 & - \\
AS-Net \citep{setprediction_hoi} & R-50 & 53.9 & - \\
QPIC \citep{qpic_hoi}  & R-50 & \underline{58.8} & 61.0 \\
SCG \citep{zhang2021spatially} & R-50 & 54.2 & 60.9 \\
MSTR \citep{kim2022mstr} & R-50 & 62.0 & 65.2 \\
SSRT \citep{iftekhar2022look}  & R-101 & 65.0 & 67.1 \\
CDN \citep{mining_hoi}  & R-101 & 63.9 & 65.9 \\
STIP \citep{zhang2022exploring} & R-50 & 66.0 & 70.7 \\
DOQ \citep{doq} & R-50 & 63.5 & - \\
UPT \citep{zhang2022efficient} & R-101 & 61.3 & 67.1 \\
IF \citep{liu2022interactiveness} & R-50 & 63.0 & 65.2\\
GEN-VLKT \citep{GEN_VLKT} & R-101 & \underline{63.6} & 65.9 \\
ViPLO \citep{ViPLO} & ViT-B/16 & 62.2 & 68.0 \\
PViC \citep{pvic} & Swin-L & \underline{64.1} & 70.2 \\
\midrule
Ours (QPIC) & R-50 & 62.4{\color{blue}$\uparrow_{3.6}$} & 64.7 \\
Ours (GEN-VLKT) & R-101 & 66.5{\color{blue}$\uparrow_{2.9}$} & 68.5 \\
Ours (PViC) & Swin-L & \textbf{66.8}{\color{blue}$\uparrow_{2.7}$} & \textbf{72.7} \\
\bottomrule
\end{tabular}
\caption{\textbf{Performance of various HOI detectors on V-COCO.} Similarly, \underline{underline} indicate the results to be compared. $AP^{S1}_{role}$ and $AP^{S2}_{role}$ represent the average precision under two different testing scenarios. Our method can achieve consistent improvements in three differently-sized HOI detectors.} 
\label{tab:sota_coco}
% \vspace{-10mm}
\end{subtable}\hfill
\begin{subtable}{.43\linewidth}
\centering
\small
\renewcommand{\arraystretch}{1.175}
\setlength{\tabcolsep}{1.5pt}
\begin{tabular}{l|ccc|ccc}
\toprule
Method & CC & KD & DE & $AP$  \\
\midrule
\multirow{6}{*}{\makecell{QPIC \\ ~\citep{qpic_hoi} \\ R-50}}   & & & & 29.07 \\
&\checkmark &            &            & 30.44{\color{blue}$\uparrow_{1.37}$} \\
&           & \checkmark &            & 30.01{\color{blue}$\uparrow_{0.94}$} \\
&           &            & \checkmark & 30.09{\color{blue}$\uparrow_{1.02}$} \\
&\checkmark & \checkmark &            & 31.26{\color{blue}$\uparrow_{2.19}$} \\
&\checkmark & \checkmark & \checkmark & 32.23{\color{blue}$\uparrow_{3.16}$} \\
\midrule
\multirow{6}{*}{\makecell{GEN-VLKT \\ ~\citep{GEN_VLKT} \\ R-101}} &&&& 34.95 \\
&\checkmark &            &            & 36.21{\color{blue}$\uparrow_{1.26}$} \\
&           & \checkmark &            & 35.83{\color{blue}$\uparrow_{0.88}$} \\
&           &            & \checkmark & 35.90{\color{blue}$\uparrow_{0.95}$} \\
&\checkmark & \checkmark &            & 36.92{\color{blue}$\uparrow_{1.97}$} \\
&\checkmark & \checkmark & \checkmark & 37.79{\color{blue}$\uparrow_{2.84}$} \\
\midrule
\multirow{6}{*}{\makecell{PViC \\ ~\citep{pvic} \\ Swin-L}}&&&        & 44.32 \\
&\checkmark &            &            & 44.99{\color{blue}$\uparrow_{0.67}$} \\
&           & \checkmark &            & 44.78{\color{blue}$\uparrow_{0.46}$} \\
&           &            & \checkmark & 44.85{\color{blue}$\uparrow_{0.53}$} \\
&\checkmark & \checkmark &            & 45.38{\color{blue}$\uparrow_{1.06}$} \\
&\checkmark & \checkmark & \checkmark & 45.71{\color{blue}$\uparrow_{1.39}$} \\
\bottomrule
\end{tabular}
\caption{\textbf{The effectiveness of the proposed methods}. We test the performance of each component of CycleHOI on the HICO-DET dataset using three different methods. \textbf{CC}: Cycle Constitency. \textbf{KD}: Knowledge Distillation. \textbf{DE}: Dataset Enhancement.}
\label{tab:module_benefit}
\end{subtable}
\caption{The performance of regular detection on the V-COCO dataset and the capability of each component on the HOI detectors of different size.}
\label{tab:cocosota_and_ablation}
\end{table}
}

\newcommand{\zeroshottable}{
\begin{table}[ht]
% \vspace{-2mm}
\small
\setlength{\tabcolsep}{5pt}
\centering
\renewcommand{\arraystretch}{1.2}
\begin{tabular}{lc|ccc}
\toprule
Method & Type & Unseen & Seen & Full \\
\midrule %%% UC
% ConsNet~\citep{consnet} & UC & 13.16 & 24.23 & 22.01 \\
HOICLIP~\citep{hoiclip} & UC & 23.15 & \textbf{31.65} & \textbf{29.93} \\
EoID~\citep{eoid} & UC & 23.01 & 30.39 & 28.91 \\
\rowcolor{gray!8}GEN-VLKT~\citep{GEN_VLKT} & UC & 20.64 & 27.16 & 25.23\\
\rowcolor{gray!8}Ours(GEN-VLKT) & UC & \textbf{23.78} & 30.07 & 28.32{\color{blue}$\uparrow_{3.09}$} \\
\hline %%% RF-UC
FCL~\citep{fcl} & RF-UC & 13.16 & 24.23 & 22.01 \\
HOICLIP~\citep{hoiclip}& RF-UC & \textbf{25.53} & 34.85 & 32.99 \\
EoID~\citep{eoid} & RF-UC & 22.04 & 31.39 & 29.52 \\
\rowcolor{gray!8}GEN-VLKT~\citep{GEN_VLKT} & RF-UC & 21.36 & 32.91 & 30.56 \\
\rowcolor{gray!8}Ours(GEN-VLKT) & RF-UC & 24.38 & \textbf{35.64} & \textbf{33.42}{\color{blue}$\uparrow_{2.86}$} \\
\hline %%% NF-UC
FCL~\citep{fcl} & NF-UC & 18.66 & 19.55 & 19.37 \\
HOICLIP~\citep{hoiclip} & NF-UC & 26.39 & \textbf{28.10} & \textbf{27.75} \\
EoID~\citep{eoid} & NF-UC & 26.77 & 26.66 & 26.69 \\
\rowcolor{gray!8}GEN-VLKT~\citep{GEN_VLKT} & NF-UC & 25.05 & 23.38 & 23.71 \\
\rowcolor{gray!8}Ours(GEN-VLKT) & NF-UC & \textbf{28.63} & 25.95 & 26.76{\color{blue}$\uparrow_{3.05}$} \\
\midrule %%% UO
% FCL~\citep{fcl} & UO & 0.00 & 13.71 & 11.43 \\
HOICLIP~\citep{hoiclip} & UO & \textbf{16.20} & 30.99 & 28.53 \\
\rowcolor{gray!8}GEN-VLKT~\citep{GEN_VLKT} & UO & 10.51 & 28.92 & 25.63 \\
\rowcolor{gray!8}Ours(GEN-VLKT) & UO & 13.92 & \textbf{32.04} & \textbf{28.86}{\color{blue}$\uparrow_{3.23}$} \\
\midrule  %%% UV
HOICLIP~\citep{hoiclip} & UV & 24.30 & 32.19 & 31.09  \\ 
EoID~\citep{eoid} & UV & 22.71 & 30.73 & 29.61 \\
\rowcolor{gray!8}GEN-VLKT~\citep{GEN_VLKT} & UV & 20.96 & 30.23 & 28.74 \\
\rowcolor{gray!8}Ours(GEN-VLKT) & UV & \textbf{24.47} & \textbf{32.83} & \textbf{31.72}{\color{blue}$\uparrow_{2.98}$} \\
\bottomrule
\end{tabular}
\caption{\textbf{Zero-shot performance comparison with state-of-the-art methods on HICO-DET.} RF is short for rare first, NF is short for non-rare first, and
UC, UO, UV indicate unseen composition, unseen object and unseen verb settings, respectively. } 
\label{tab:zero_shot}
% \vspace{-5mm}
\end{table}
}

\newcommand{\ablationtable}{
\begin{table}[ht]
\centering
\small
\setlength{\tabcolsep}{2.5pt}
\begin{subtable}{.33\linewidth}  
\begin{tabular}{c|ccc}
\makecell{feature \\ map} & Full & Rare & \makecell{Non- \\ Rare} \\
\hline
None   & 29.07 & 21.85 & 31.23 \\
Stage0 & 29.59{\color{blue}$\uparrow_{0.52}$} & 22.32 & 31.76 \\
Stage1 & 29.72{\color{blue}$\uparrow_{0.65}$} & 22.48 & 31.91 \\
Stage2 & 29.86{\color{blue}$\uparrow_{0.79}$} & 22.67 & 32.13 \\ \rowcolor{gray!15}
Stage3 & 30.12{\color{blue}$\uparrow_{1.05}$} & 22.84 & 32.29 \\
All    & 29.74{\color{blue}$\uparrow_{0.67}$} & 22.52 & 31.97 \\
\end{tabular}
\caption{\textbf{U-Net feature map} \\ for knowledge distillation.}  
\label{tab:ablation:distill_size}  
\end{subtable}\hfill 
\begin{subtable}{.33\linewidth}  
\begin{tabular}{c|ccc}
\makecell{time \\ step} & Full & Rare & \makecell{Non- \\ Rare} \\
\hline
None & 29.07 & 21.85 & 31.23 \\
0    & 29.83{\color{blue}$\uparrow_{0.76}$} & 22.65 & 32.07 \\ \rowcolor{gray!15}
1    & 30.12{\color{blue}$\uparrow_{1.05}$} & 22.84 & 32.29 \\
10   & 29.87{\color{blue}$\uparrow_{0.80}$} & 22.58 & 32.04 \\
100  & 29.56{\color{blue}$\uparrow_{0.49}$} & 22.36 & 31.73 \\ 
500  & 29.33{\color{blue}$\uparrow_{0.26}$} & 22.13 & 31.54 \\
\end{tabular}
\caption{\textbf{Time Step} to get the \\ U-Net feature map.}  
\label{tab:ablation:distill_time}  
\end{subtable}\hfill
\begin{subtable}{.33\linewidth}  
\begin{tabular}{c|ccc}
threshold & Full & Rare & \makecell{Non- \\ Rare} \\
\hline
None & 29.07 & 21.85 & 31.23 \\
0.5 & 29.50{\color{blue}$\uparrow_{0.43}$} & 22.25 & 31.69 \\ \rowcolor{gray!15}
1.0 & 29.74{\color{blue}$\uparrow_{0.67}$} & 22.53 & 31.87 \\
1.5 & 29.59{\color{blue}$\uparrow_{0.52}$} & 22.34 & 31.80 \\
2.0 & 29.55{\color{blue}$\uparrow_{0.48}$} & 22.32 & 31.72 \\
2.5 & 29.47{\color{blue}$\uparrow_{0.40}$} & 22.27 & 31.59 \\
\end{tabular}
\caption{\textbf{Threshold} used for filtering labels.}  
\label{tab:ablation:label_thresh}  
\end{subtable}

\begin{subtable}{.32\linewidth}  
\begin{tabular}{c|ccc}
\makecell{loss \\ type} & Full & Rare & \makecell{Non- \\ Rare} \\
\hline
None & 29.07 & 21.85 & 31.23 \\ 
L1   & 30.39{\color{blue}$\uparrow_{1.32}$} & 23.31 & 32.53 \\ \rowcolor{gray!15}
L2   & 30.44{\color{blue}$\uparrow_{1.37}$} & 23.24 & 32.56 \\
PL   & 29.96{\color{blue}$\uparrow_{0.89}$} & 22.76 & 32.11
\end{tabular}
\caption{Types of \textbf{Cycle Consist- \\ ency Loss}.}  
\label{tab:ablation:ti_loss}  
\end{subtable}\hfill
\begin{subtable}{.32\linewidth}  
\begin{tabular}{c|ccc}
method & Full & Rare & \makecell{Non- \\ Rare} \\
\hline
None & 29.07 & 21.85 & 31.23 \\  \rowcolor{gray!15}
M1   & 30.44{\color{blue}$\uparrow_{1.37}$} & 23.24 & 32.56 \\
M2   & 30.31{\color{blue}$\uparrow_{1.24}$} & 23.13 & 32.45 \\
M3   & 30.36{\color{blue}$\uparrow_{1.29}$} & 23.20 & 32.46
\end{tabular}
\caption{Calculation Method of \\ \textbf{Cycle Consistency Loss}.}  
\label{tab:ablation:ti_method}  
\end{subtable}\hfill
\begin{subtable}{.35\linewidth}  
\begin{tabular}{c|ccc}
method & Full & Rare & \makecell{Non- \\ Rare} \\
\hline
None & 29.07 & 21.85 & 31.23 \\
TI & 29.54{\color{blue}$\uparrow_{0.47}$} & 22.57{\color{blue}$\uparrow_{0.72}$} & 31.53 \\  \rowcolor{gray!15}
DB & 29.61{\color{blue}$\uparrow_{0.54}$} & 22.68{\color{blue}$\uparrow_{0.83}$} & 31.54 \\
\end{tabular}
\caption{\textbf{Method} used to solve long-tail problems.}  
\label{tab:ablation:long_tail}  
\end{subtable}
% \vspace{-2mm}
\caption{\textbf{Ablations}. We conduct studies on HICO-DET based on QPIC with R-50 as backbone. The best setting is marked gray.}
\label{tab:ablations}
% \vspace{-5mm}
\end{table}
}

\newcommand{\traincosttable}{
\begin{wraptable}{l}{65mm}
% \vspace{-8mm}
\centering
\scriptsize
\setlength{\tabcolsep}{1pt}
\renewcommand{\arraystretch}{1.2}
\begin{tabular}{cc|ccc|c}  
Method & Backbone & mem. & time & GFLOPs & mAP\\  
\hline  
\multirow{4}{*}{\makecell{QPIC\\(baseline)}}& R-50 & 14GB & 22m/ep & 226 & 29.07\\ 
& SwinT & 17GB & 34m/ep & 247 & 30.87\\   
& \cellcolor{gray!25}SwinB & \cellcolor{gray!25}22GB & \cellcolor{gray!25}47m/ep & \cellcolor{gray!25}539 & \cellcolor{gray!25}32.58\\
& SwinL & 31GB & 62m/ep & 921 & 35.43\\
\hline
QPIC+ours & \cellcolor{gray!25}R-50 & \cellcolor{gray!25}23GB & \cellcolor{gray!25}49m/ep & \cellcolor{gray!25}557 & \cellcolor{gray!25}32.23\\
\end{tabular}
% \vspace{-2mm}
\captionof{table}{\textbf{Training consumption and performance}. \textbf{mem.}: GPU memory. \textbf{time}: training time, minutes per epoch.}
\label{tab:train_cost}
% \vspace{-6mm}
\end{wraptable} 
}
\newcommand{\datasetissuefigure}{
\begin{figure}[ht]
\centering
\begin{subfigure}{0.531\linewidth}
\includegraphics[width=\linewidth]{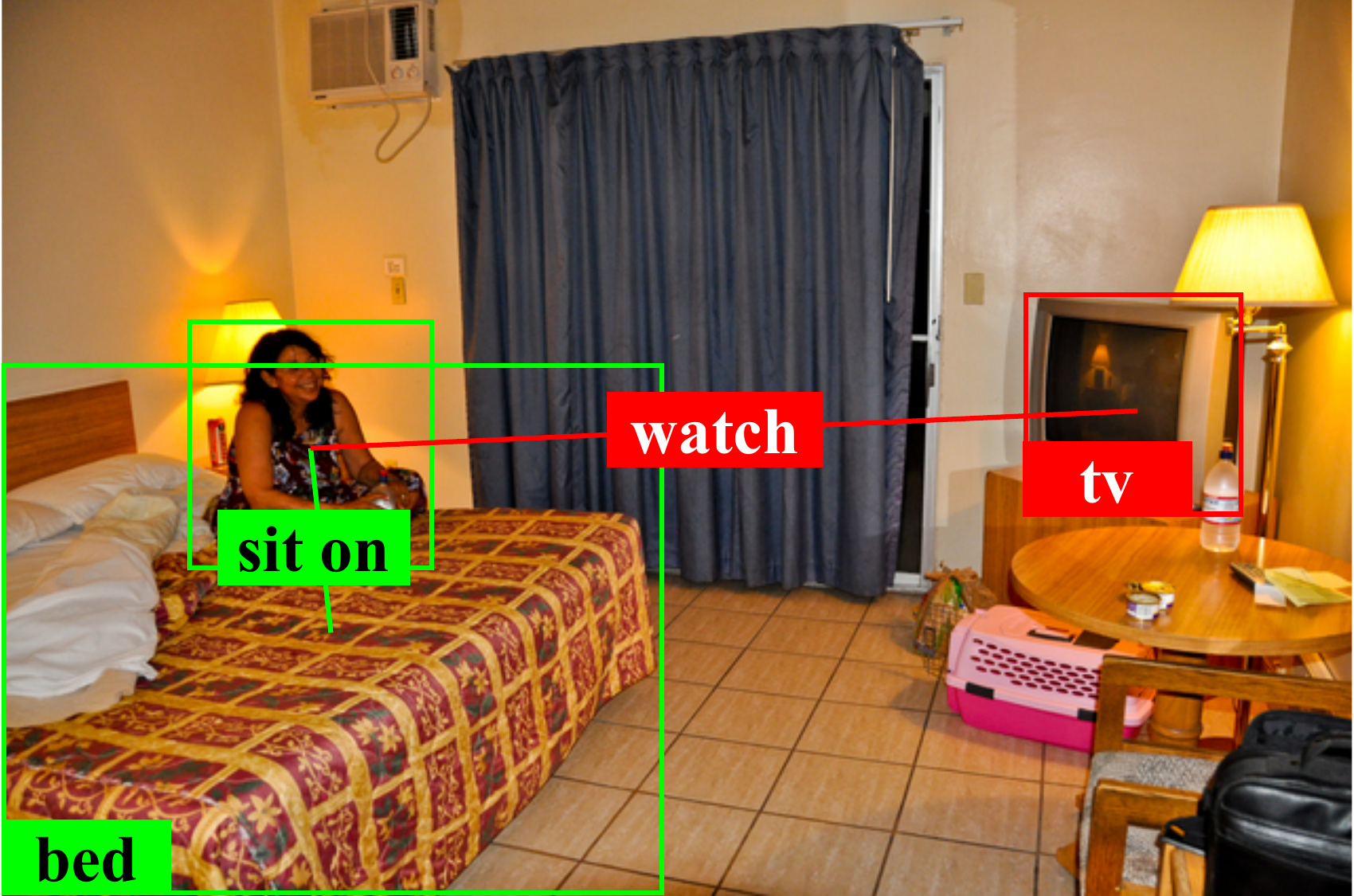}
\caption{}
\label{fig:motivation_many_rel}
\end{subfigure}
\begin{subfigure}{0.24\linewidth}
\includegraphics[width=\linewidth]{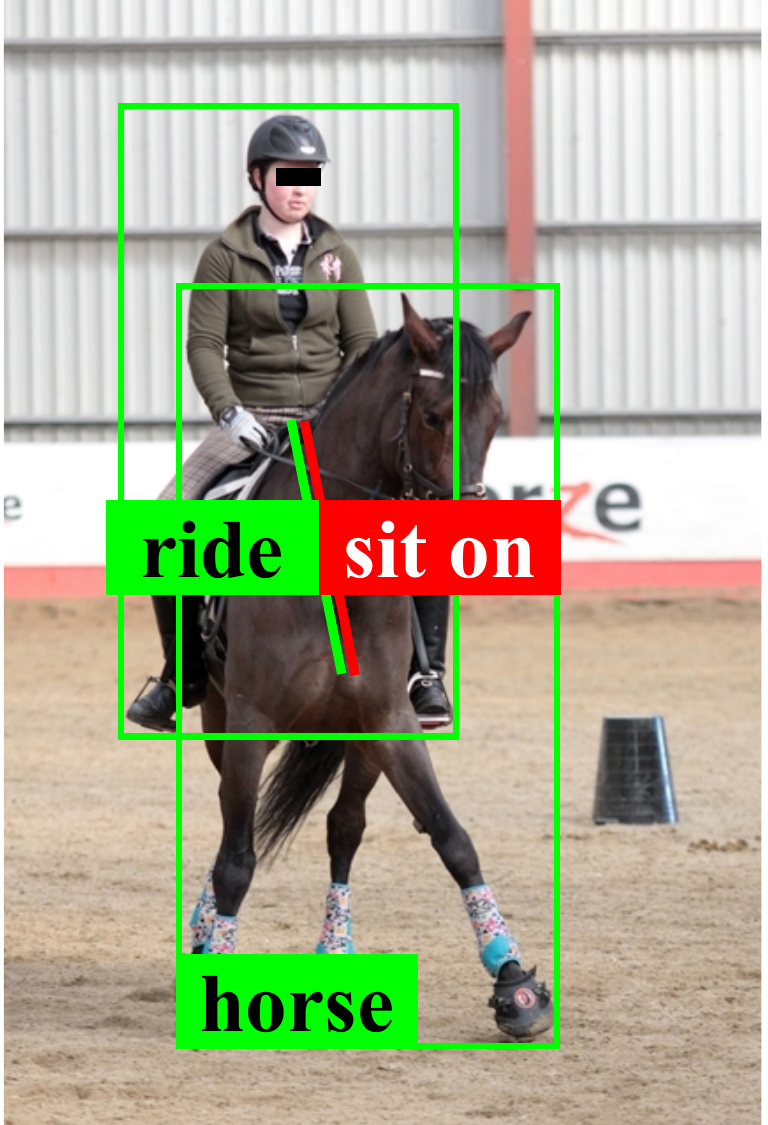}
\caption{}
\label{fig:motivation_lack_obj}
\end{subfigure}\hfill
\begin{subfigure}{0.4\linewidth}
\includegraphics[width=\linewidth]{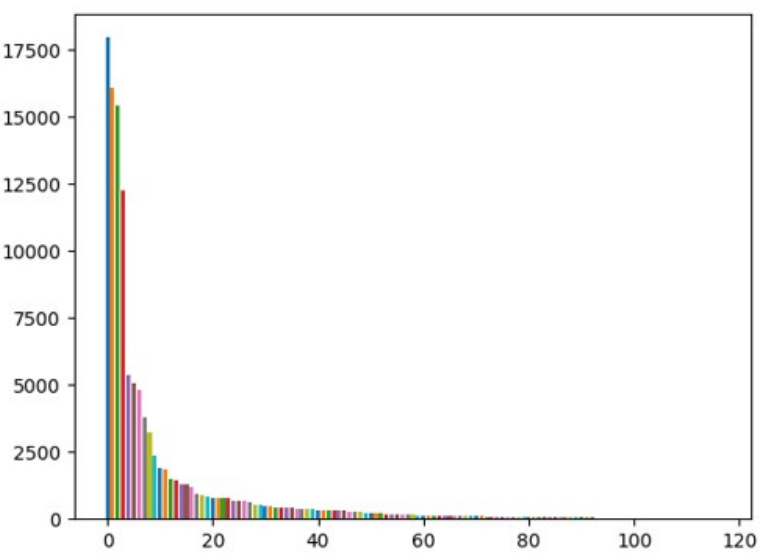}
\caption{HICO-DET}
\label{fig:motivation_hico_longtail}
\end{subfigure}
\begin{subfigure}{0.4\linewidth}
\includegraphics[width=\linewidth]{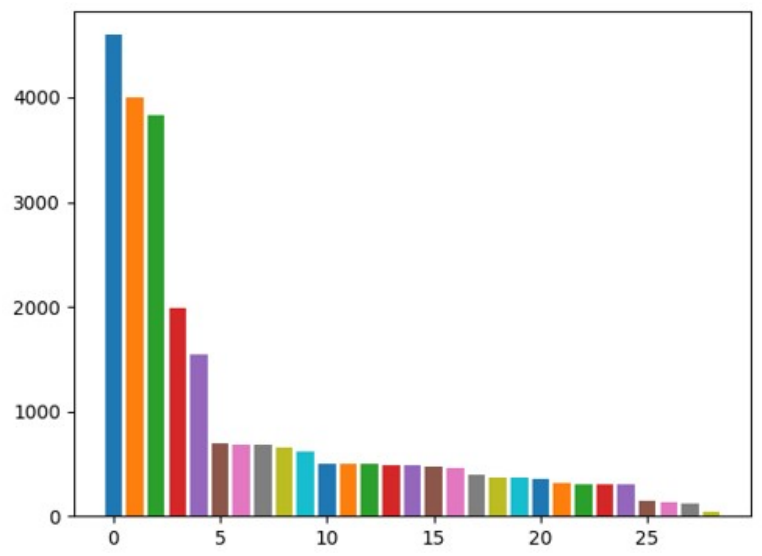}
\caption{VCOCO}
\label{fig:motivation_vcoco_longtail}
\end{subfigure}
% \vspace{-0.3cm}
\caption{The critical issues in the HOI datasets. (a) The triplet $\langle human, ride, horse \rangle $ is annotated, but the relation $sit~on$ is neglected. (b) The triplet $\langle human, sit~on, bed \rangle $ is annotated, but $\langle human, watch, TV \rangle $ is neglected. The ground-truth triplets are marked in green with black texts. The missing objects or relations are marked in red with white texts. (c) and (d) The extreme long-tailed relation distribution in the existing HOI datasets.}
% \vspace{-5mm}
\label{fig:motivation}
\end{figure}
}

\newcommand{\frameworkfigure}{
\begin{figure}[ht]
% \vspace{-5mm}
\centering
\includegraphics[width=\linewidth]{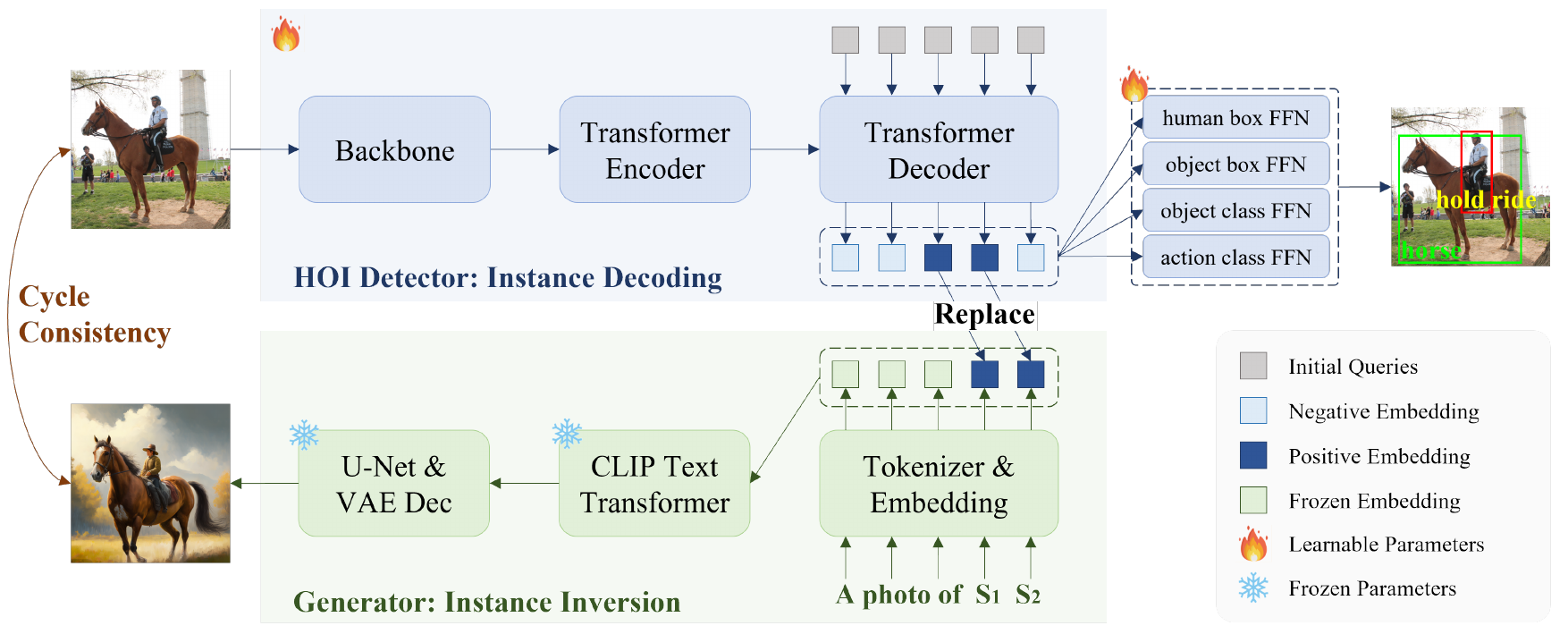}
\caption{{\bf Pipeline of CycleHOI}. We propose an enhanced training framework for improving the generalization ability of learned HOI detector, which is composed of an instance decoding process and an instance inversion process. The detection-generation cycle consistency loss is applied on top of two processes to enforce them to be compatible with each other. This new cycle consistency design allows us to bridge the pre-trained diffusion models with the DETR-alike HOI detection pipeline, thereby improving its performance.}
\label{fig:framework}
% \vspace{-5mm}
\end{figure}
}

\newcommand{\attentionmapfigure}{
\begin{figure}[ht]
% \vspace{-5mm}
\centering
\begin{subfigure}{0.8\linewidth}
\includegraphics[width=\linewidth]{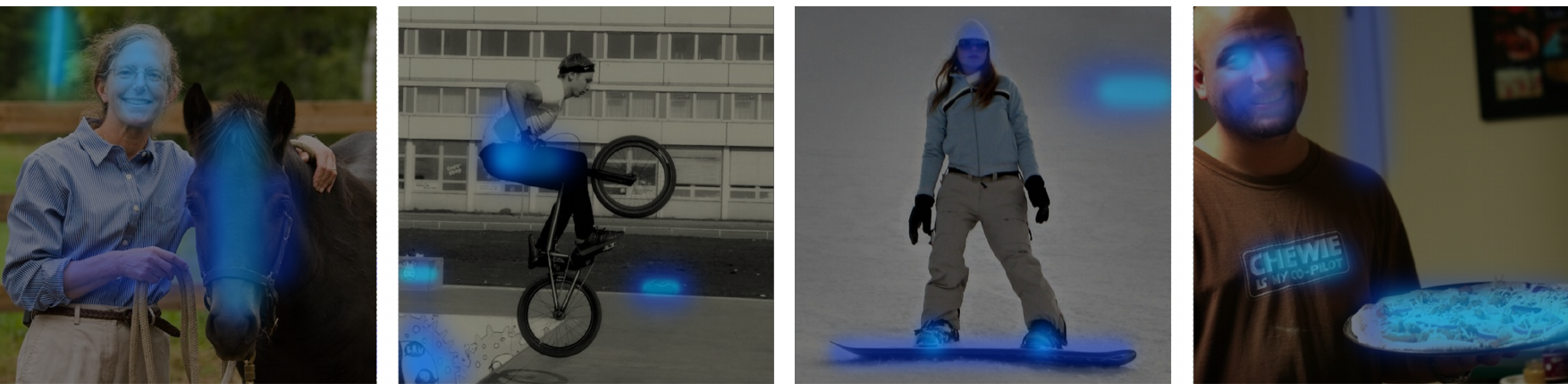}
\caption{QPIC Transformer Encoder.}
\label{fig:attn_map1}
\end{subfigure}\hfill
\begin{subfigure}{0.8\linewidth}
\includegraphics[width=\linewidth]{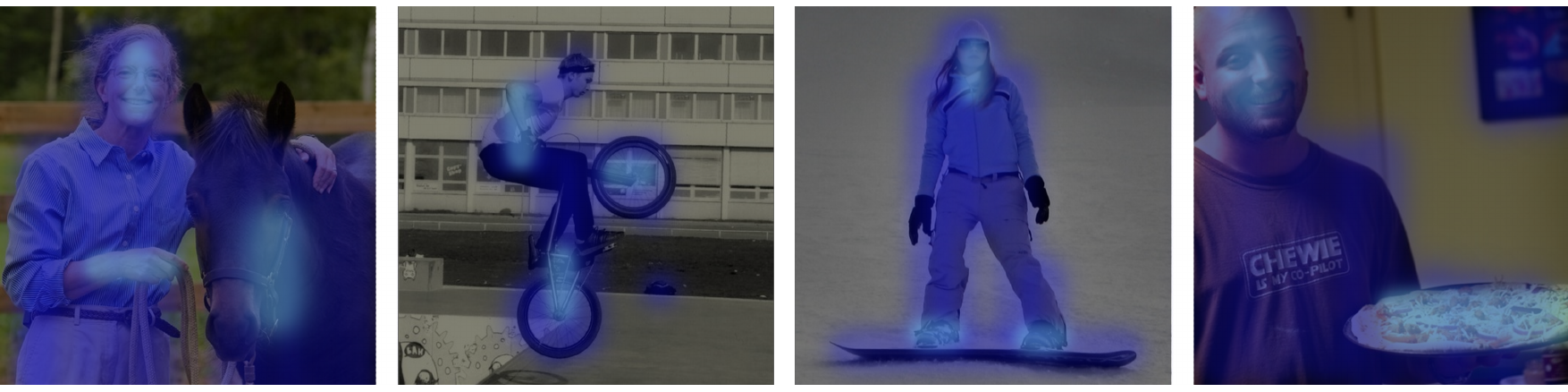}
\caption{Stable Diffusion U-Net.}
\label{fig:attn_map2}
\end{subfigure}
% \vspace{-2mm}
\caption{Visualization of attention maps of two models.}
\label{fig:attn_map}
% \vspace{-10mm}
\end{figure}
}

\newcommand{\submethodfigure}{
\begin{figure}[ht]
% \vspace{-5mm}
\centering
\begin{subfigure}{0.38\linewidth}
\includegraphics[width=\linewidth]{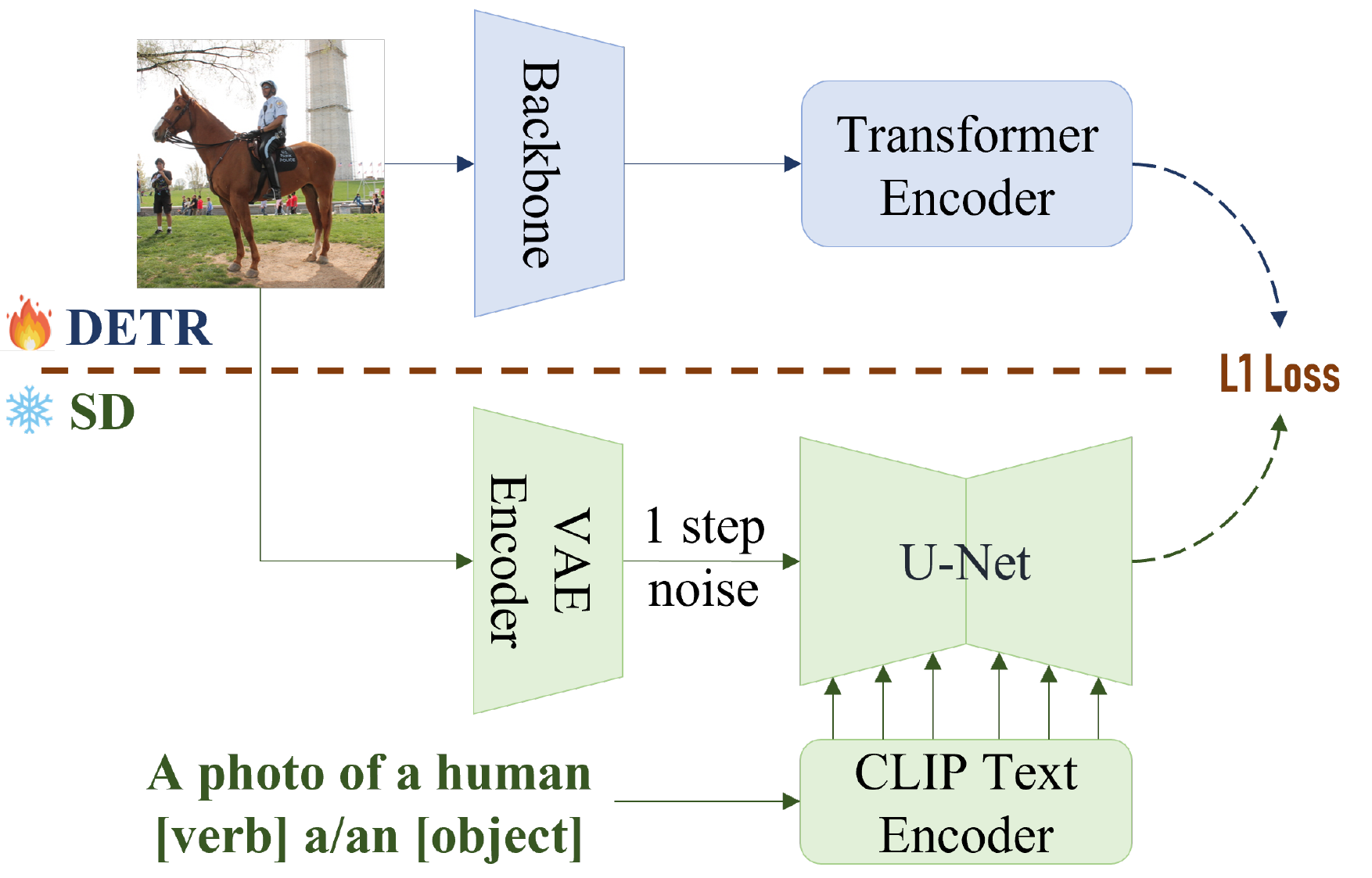}
\caption{\textbf{Knowledge distillation from diffusion model to HOI detector}. We design an one-step denoising process for diffusion model to guide the HOI detector training.}
\label{fig:method_distill}
\end{subfigure}\hfill
\begin{subfigure}{0.59\linewidth}
\includegraphics[width=\linewidth]{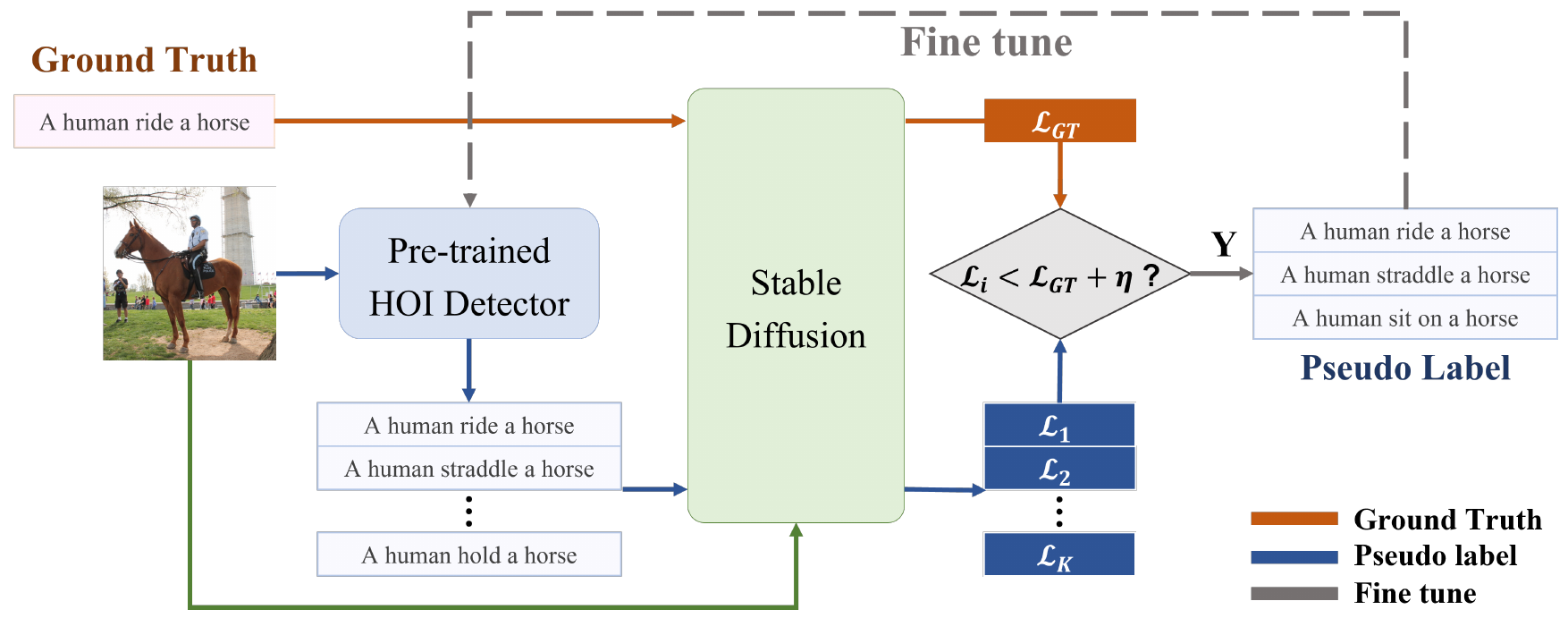}
\caption{\textbf{Pseudo-label generation process}. We treat Stable Diffusion as an external knowledge base, and use its loss to determine which of the predictions from the pre-trained HOI detectors qualify as pseudo-labels. These pseudo-labels are later used to fine-tune the HOI detector.}
\label{fig:method_pseudo}
\end{subfigure}
% \vspace{-2mm}
\caption{Knowledge distillation and Pseudo-label generation.}
\label{fig:distill_pseudo}
% \vspace{-10mm}
\end{figure}
}

\begin{abstract}
\label{sec:abstruct}
Recognition and generation are two fundamental tasks in computer vision, which are often investigated separately in the exiting literature. 
However, these two tasks are highly correlated in essence as they both require understanding the underline semantics of visual concepts. 
In this paper, we propose a new learning framework, coined as CycleHOI, to boost the performance of human-object interaction (HOI) detection by bridging the DETR-based detection pipeline and the pre-trained text-to-image diffusion model. Our key design is to introduce a novel cycle consistency loss for the training of HOI detector, which is able to explicitly leverage the knowledge captured in the powerful diffusion model to guide the HOI detector training. Specifically, we build an extra generation task on top of the decoded instance representations from HOI detector to enforce a detection-generation cycle consistency. Moreover, we perform feature distillation from diffusion model to detector encoder to enhance its representation power. In addition, we further utilize the generation power of diffusion model to augment the training set in both aspects of label correction and sample generation. We perform extensive experiments to verify the effectiveness and generalization power of our CycleHOI with three HOI detection frameworks on two public datasets: HICO-DET and V-COCO. The experimental results demonstrate our CycleHOI can significantly improve the performance of the state-of-the-art HOI detectors. 
\end{abstract}  
\section{Introduction}
\label{sec:intro}

Human-object interaction (HOI) detection~\citep{hotr,setprediction_hoi,qpic_hoi,e2e_hoi,hoi_nlp_prior,mining_hoi} aims to detect humans and the corresponding interaction objects with their pairwise relations in an image.
In contrast to the normal relation detection tasks like scene graph generation~\citep{trace,structured_sparse_rcnn,dynamic_sgg,neural_motifs,tang_treelstm,unbias_sgg,prior_vrd,gu_sgg_gan}, HOI~\citep{hicodet,vcoco} focuses on the human actions involving objects, such as \textit{carrying} and \textit{holding}, without consideration of spatial relation labels.
Current HOI detection methods often follow the similar training paradigm of 2D object detectors~\citep{detr,sparsercnn,adamixer,stageinteractor,deformabledetr}, and use the $\langle human, verb, object \rangle $ triplet annotations from the existing datasets to supervise the triplet predictions.
\datasetissuefigure

Unlike the traditional object detection, HOI involves the complex reasoning over the relation between human and interaction objects, which poses challenges to build a high-quality HOI dataset. 
First, it is almost impossible for annotators to label all possible relations under limited labors. This is because the relations are diverse in the real world and hard to precisely define~\citep{hoi_nlp_prior}.
For example, there is a \textit{human} riding on a \textit{horse} in Figure~\ref{fig:motivation_many_rel}. Although the annotators have been aware of the existence of the relation \textit{riding}, other missing relations such as \textit{sitting on} are also plausible here.
Second, some relations are ubiquitous but easily overlooked by human annotators. As depicted in Figure~\ref{fig:motivation_lack_obj}, the triplet $\langle human, watch, TV \rangle $ is totally neglected by the annotators.
Finally, these relation categories often exhibit a long-tail distribution~\citep{mining_hoi}. Figure~\ref{fig:motivation_hico_longtail} and~\ref{fig:motivation_vcoco_longtail} illustrate the relation label distribution in the HICO-DET~\citep{hicodet} and VCOCO~\citep{vcoco} datasets, respectively. In the HICO-DET dataset, the top relation categories have more than 15,000 images, while several tail relation categories only have as few as one image, such as ``zip" and ``flush". These critical issues make it very difficult to train an effective HOI detector solely on the existing datasets.

Recently the text-to-image diffusion models~\citep{ddpm, ddpm2, ddim, stable_diffusion} have achieved tremendous success in the field of generation and are able to produce high-quality images. This is attributed to its meticulously designed extensive network architecture and a vast amount of training data. Although generation and detection are two different tasks and often investigated separately in the exist works, we argue that {\em they are highly correlated as they both require understanding the underline semantics of visual concepts}. So, a natural question arises {\em whether we can leverage the pre-trained diffusion models to assist the training of HOI detector to mitigate the above issues}? Intuitively, these pre-trained diffusion models have already captured the rich knowledge about visual concepts, which is expected to be helpful to improve the generalization ability of HOI detector trained from these weakly-annotated and challenging HOI datasets.

Based on the above analysis, in this paper, we propose an enhanced training framework for HOI detectors via bridging the DETR-based HOI detection pipeline and the pre-trained text-to-image diffusion models. To relieve the training difficulty on the weakly-annotated HOI dataset, our key design is to introduce a novel cycle consistency constraint on the detected HOI instances. Our basic idea is to couple the instance decoding process in the DETR-alike HOI detector with an instance inversion process to reconstruct the image from the detection results. In this sense, we introduce the detection-generation cycle consistency loss to encourage the decoded instance to keep the key information for re-generating the original image. Specifically, we use a pre-trained text-to-image diffusion model and replace the corresponding input text embeddings with our decoded HOI query features. Then, the updated text inputs are passed through the pre-trained text-to-image diffusion model to invert the decoded instances to the reconstructed image, which is enforced to be similar to the original image. This cycle consistency constraint yields a natural bridge to connect the HOI detection pipeline to the text-to-image diffusion models.

In addition, to further enhance its representation ability, we design a simple knowledge distillation strategy from diffusion models to DETR-alike detector by explicitly building an one-step de-noising process. We minimize the feature difference between the DETR encoder and the U-Net from the diffusion model. This distillation process is able to enable representations to attend on more diverse and discriminative regions. Moreover, from a more practical view, we treat the diffusion model as an additional knowledge repository for correcting the dataset by generating missing labels and augmenting images of rare categories. We employ the loss from the diffusion model to filter HOI detection predictions, and use them as pseudo-labels to address dataset label omissions. We utilize DreamBooth~\citep{dreambooth} to learn personalized concepts in rare categories, thus generating images with similar concepts to tackle the long-tail problem. We perform experiments on two HOI detection datasets: HICO-DET~\citep{hicodet} and V-COCO~\citep{vcoco}. Experiment results demonstrate that our proposed method yields significant improvements across various HOI detectors. In addition, we perform detailed ablation studies to show the effectiveness of our proposed designs. In summary, our main contribution is threefold:

\begin{itemize}
\item We introduce a new cycle consistency constraint on the HOI detector training via bridging the detection pipeline and the pre-trained diffusion models. Our cycle consistency is a general design and could be applied to any HOI detector to improve its performance without introducing any extra cost in inference phase.
\item We further explore complementary ways to exploit diffusion models to enhance feature representation and augment training set. These simple yet practical strategies turns out to be effective to mitigate the common issues within the existing HOI datasets.
\item The experiment results demonstrate that our proposed CycleHOI can significantly improve the performance of multiple HOI detectors. Additionally, we offer in-depth ablation studies to investigate the effectiveness of our proposed methods.
\end{itemize}

\section{Related Work}
\label{sec:relate_work}

\subsection{Human-Object Interaction Detection} 
\label{sec:related_HOI}
Human-object interaction (HOI) detection is a task that requires a detector to localize and recognize each human-object pair in an image and predict the semantic relation in each pair.
There are many methods which first detect all the humans and objects in an image, and then pair them and classify the interactions~\citep{hoi_analysis,hoi_two_stage1}.
Several HOI detectors~\citep{ppdm,zhong2021glance,wang2020learning} aim to detect humans, objects and interactions at the same time.
These detectors typically use two branches to perform instance detection and interaction detection in parallel, and a matching algorithm is used to fuse the outputs of these two branches.
Since DETR~\citep{detr} was proposed, plenty of works about query-based HOI detectors have been proposed~\citep{hotr,setprediction_hoi,qpic_hoi,mining_hoi,GEN_VLKT}.
These query-based HOI detectors also belong to the one-stage detector. They are usually based on a set of learnable triplet queries which progressively aggregate the features through cascade decoder layers. The outputs of the final decoder layers are the HOI predictions.
Based on the query-based object detection paradigm, there are also plenty of works enhance the quality of detection with knowledge distillation~\citep{doq}, priors from CLIP~\citep{hoiclip,GEN_VLKT} or natural language prior~\citep{hoi_nlp_prior}.
In this paper, the method we propose can be applied to any HOI detector based on DETR, serving as a plug-and-play approach to assist in the training of HOI detectors. During inference, all the methods we introduce can be removed, returning the HOI detector to its pure state.

\subsection{Diffusion Models}
\label{sec:related_diffusion}
Recently, diffusion models~\citep{ddpm,ddpm2,ddim,beats_gan,scorebase,sliced_scorebase,edm,cold_diffusion,unified_perspective_diffusion,analytic_DPM} have achieved great success in text-to-image generation~\citep{stable_diffusion,imagen,dalle,attribute_t2i}.
The diffusion model consists of two processes: a forward process where images are gradually corrupted into Gaussian noises, and a backward process where the images are restored by using a learnable denoising network.
The vanilla diffusion models typically require thousands of backward steps to generate an image from the pure Gaussian noise. To accelerate this process, several methods about introducing training-free samplers~\citep{dpm_solver,dpm_solver2} or knowledge distillation~\citep{diffusion_pd,consistency_model} have been proposed. 
There are also several applications of pre-trained diffusion models.
For example, Textual Inversion~\citep{textual_inversion} uses pre-trained diffusion models to represent user-provided concepts (\emph{e.g.}, attributes, objects or even relations~\citep{textual_inversion_relation_gen_image}) with learned word embeddings, and the model can generate relevant images according to these new embeddings. 
DreamBooth~\citep{dreambooth} is another type of generative model that learns personalized concepts. Unlike textual inversion, which learns word embeddings, it fine-tunes the network directly based on user input images and personalized concept prompts. To prevent overfitting to personalized concepts, an additional pre-trained diffusion model is used in a frozen state to supervise it, without the addition of personalized concept prompts.
The pre-trained diffusion models can also perform zero-shot image classification to some extent~\citep{diffusion_cls1,diffusion_cls2}, and we can enhance its discriminability by setting additional classification supervision~\citep{egc}.
In this paper, we propose to apply pre-trained text-to-image generative models to aid the training of HOI detection.

\section{Method}
\label{sec:method}
\subsection{Preliminaries}
\label{section: preliminaries}
\textbf{Text-to-Image Diffusion Model}. Diffusion model is a type of generative model that progressively transforms a Gaussian noise $\mathbf{x}_T$ into a meaningful image $\mathbf{x}_0$. During training, an image or its latent representation is corrupted by a Gaussian noise, and then the network in the diffusion model learns to restore it. The diffusion model is able to generate specific images given texts with the help of an additional pre-trained text encoder. The reconstruction loss for training text-to-image diffusion models is defined as follows:
\begin{equation}
\label{eq:reconstruction}
\mathcal{L} = \mathbb{E}_{\mathcal{E}(\mathbf{x}), y, \epsilon \sim \mathcal{N}(0,1), t}[ \Vert \epsilon - \epsilon_\theta (z_t, t, \tau_\theta(y)) \Vert^2_2 ],
\end{equation}
where $\epsilon$ denotes a Gaussian noise. $t$ denotes the timestep ranging from $0$ to $T$. $\epsilon_\theta (\cdot,\cdot,\cdot)$ denotes a denoising network which is typically a time-conditional U-Net~\citep{unet}.
$z_t$ denotes the image in latent space and is obtained from $\mathcal{E}$. $y$ denotes a condition such as the textual prompt ``\textit{a photo of ...}''. $\tau_\theta(\cdot)$ is a encoder which encodes the condition $y$.
In Stable Diffusion, $\tau_\theta(\cdot)$ is a CLIP~\citep{clip} text encoder, where each word in a textual prompt is mapped to an embedding.
\frameworkfigure
\subsection{CycleHOI}
\label{sec:overview}
To improve the performance of the learned HOI detector, we propose an enhanced training framework, termed as {\em CycleHOI} as shown in Figure~\ref{fig:framework}, by bridging the DETR-based HOI detection pipeline and the pre-trained text-to-image diffusion model through a cycle-consistency constraint. We will give a detailed description on this CycleHOI training framework in this section. In addition, to further enhance its representation ability of HOI detector, we devise a knowledge distillation strategy from the pre-trained text-to-image diffusion model to its DETR encoder as shown in Figure~\ref{fig:method_distill} via an one-step denoising process. This knowledge distillation strategy is able to guide the transformer encoder to attend on more diverse and discriminative regions of image, thus leading to a better detection performance. 

\subsubsection{Detection and Generation Cycle Consistency.}
\label{sec:cycle}

Our CycleHOI training framework is composed of two processes: instance decoding process and instance inversion process. The instance decoding process is a normal HOI detector based on the DETR framework~\citep{detr}, which is composed of a backbone, a transformer encoder, and a transformer decoder. The instance inversion process is modified image-to-text diffusion model, where the input text embeddings are replaced with HOI detector query vectors for the original image reconstruction. The detection-generation cycle consistency is applied on both process to enforce them to be compatible with each other.

{\bf HOI Detector: Instance Decoding Process.} Our HOI detector baseline chooses a DETR-alike detection pipeline, and in experiments it could be QPIC~\citep{qpic_hoi}, GEN-VLKT~\citep{GEN_VLKT}, or PViC~\citep{pvic}. Formally, an image $\boldsymbol{I}$ is first fed into a backbone and a transformer encoder to form a feature map $\boldsymbol{F}$. Then, in the transformer decoder, some initialized queries $\boldsymbol{Q}$ go through self-attention and perform cross-attention with the feature map to decode the HOI instances from the image content. Finally, these updated queries are fed into FFNs to directly predict the human box, object box, object class and action class. The original training of DETR-alike detection pipeline is based on the bipartite graph matching between the query vectors and the ground-truth. During this matching process, some queries are assigned to the foreground action instances while the other queries are assigned as the background class. Based on this optimal matching, the standard detection loss $\mathcal{L}_{Det}$, as demonstrated in Eq. 6 in \citep{qpic_hoi}, including cross entropy loss for object classification, focal loss~\citep{focalloss} for relation classification, the L1 and GIoU loss for human and object bounding box regression are applied to guide the HOI detector training.

{\bf Generator: Instance Inversion Process.} Inspired by Textual Inversion~\citep{textual_inversion}, our image generator from the decoded instance representation is based on a modified text-to-image diffusion model. Its objective is to reconstruct the image from the decoded instance representation and couple the decoding process with inversion process. Formally, we choose a standard pre-trained text-to-image diffusion model~\citep{stable_diffusion}. According to the HOI ground-truth, we build a text prompt of ``{\em a photo of $S_*$}'', where $S_*$ is the special token to represent the corresponding HOI class. Then, this text prompt is passed through a tokenizer to generate the word embeddings. Finally, we replace the special embedding of $S_*$ with the corresponding positive queries determined by the bipartite graph matching. These updated text embeddings will be fed into a pre-trained diffusion model to reconstruct the image. Specifically, we categorize the embeddings output by the Transformer Decoder of the detector into two types: positive embeddings and negative embeddings. Embeddings that match with the ground-truth during the detector's bipartite graph matching process are called positive embeddings; otherwise, they are negative embeddings. Suppose the current image contains $m$ HOI (Human-Object Interaction) instances, which means there are $m$ ground-truth annotations, then there would be $m$ positive embeddings. At this time, the text prompt becomes "A photo of $S_1$, $S_2$, ..., $S_m$," where each $S_i$, for $i=1, 2, ..., m$, represents a specific HOI instance corresponding to the $m$ annotations. For each annotation, or $S_i$, we replace the embedding generated by $S_i$ in the text prompt with the corresponding positive embedding. It is important to note that the number of HOI categories in the dataset is equal to the number of different $S$ types. Here, $S$ does not play an actual role and is not involved in the training of the network. It is merely used to distinguish which HOI category each positive embedding specifically corresponds to, thus facilitating targeted optimization by the diffusion model in subsequent stages.

Based on the two processes of instance decoding and instance inversion, we build our CycleHOI training framework by applying a consistency loss between them. Intuitively, we hope the decoded HOI instances should convey enough information about the image content and we are able to reconstruct the original image based on the detection results. Formally, we design the following detection-generation cycle consistency loss:
\begin{equation}
    \mathcal{L}_{Cycle} = \|\mathrm{Gen}(\mathrm{HOIDet}(\boldsymbol{I})) - \boldsymbol{I}\|_2,
\end{equation}
where $\mathrm{HOIDet}$ represents the HOI detector and $\mathrm{Gen}$ represents the generator. It should be noted that during the training process, the diffusion model is frozen and we only focus on optimizing the parameters of HOI detector.
\attentionmapfigure

\subsubsection{Feature Distillation From Diffusion Model.}
\label{sec:distill}

In order to improve the feature representation power of HOI detector, we propose to distill knowledge from a pre-trained text-to-image diffusion model. 
The diffusion model is typically trained on huge numbers of images with a large model capacity. So, we expect this large-scale pre-trained diffusion model can capture more effective representation of the visual concepts. We perform a visual comparison of the pre-trained Stable Diffusion~\citep{stable_diffusion} as well as the DETR-based HOI detector QPIC~\citep{qpic_hoi} in Figure~\ref{fig:attn_map}. From the visualization results, we see that Stable Diffusion pays more attention to people and objects than QPIC. Therefore, we use Stable Diffusion as a teacher network to guide the training of HOI detector and transfer the knowledge rich in the diffusion model to the HOI detector, as shown in Figure~\ref{fig:method_distill}.

Specifically, to distill the knowledge from the text-to-image diffusion model to the HOI detector, we build an one-step denoising process. We add random Gaussian noise to the input image and input noisy image into the pre-trained diffusion model to mimic a denoising process. At the same time, we compose the ground-truth corresponding to the image into a textual prompt ``\textit{a photo of a human} [\textit{verb}] \textit{a/an} [\textit{object}]'', where [\textit{verb}] and [\textit{object}] can be filled by the categories of a relation and an object, respectively. If there are multiple pairs of human-object relations in an image, we connect the prompts by ``,'' into a long sentence. We then feed this textual prompt into the text encoder in the Stable Diffusion. Through the denoising process of one forward propagation of Stable Diffusion, we can get the output feature map $\boldsymbol{F}_S$ of U-Net. Meanwhile, through one forward propagation of HOI detector, we can also get the output feature map $\boldsymbol{F}_D$ of transformer encoder. We align the two features by down-sampling the U-Net output feature map and using a $1 \times 1$ convolution operation. Distillation is achieved by calculating the difference between these two features as follows:
\begin{equation}
\label{eq:distill}
\mathcal{L}_{Dis} = \|\boldsymbol{F}_S - \boldsymbol{F}_D\|_1.
\end{equation}

In summary, the overall loss function for our CycleHOI training framework is shown below:
\begin{equation}
\mathcal{L} = \lambda_{Det}\mathcal{L}_{Det} + \lambda_{Cycle}\mathcal{L}_{Cycle} + \lambda_{Dis}\mathcal{L}_{Dis},
\end{equation}
where $\mathcal{L}_{Det}$, $\mathcal{L}_{Cycle}$, and $\mathcal{L}_{Dis}$ denote the loss of the HOI detector, the loss of cycle consistency, and the loss of knowledge distillation, respectively. $\lambda_{Det}$, $\lambda_{Cycle}$, and $\lambda_{Dis}$ are used to adjust the weights of each loss.
\submethodfigure
\subsection{Dataset Enhancement with Diffusion Model}
\label{sec:dataset}
\textbf{Label Generation.}
As mentioned above, HOI datasets have a serious problem of missing labels. Thus, we propose an automatic way to augment the labels of training set, as shown in Figure~\ref{fig:method_pseudo}. Specifically, for each image, we use the textual prompt ``\textit{a photo of a human} [\textit{verb}] \textit{a/an} [\textit{object}]'', where [\textit{verb}] and [\textit{object}] can be filled by the HOI categories. Similarly, if there are multiple annotations, we use ``,'' to join them into a long sentence.
Then, we feed each image and its corresponding prompt into the diffusion model, so we can compute a reconstruction loss $\mathcal{L}_{GT}$ for each image. We assign the value $\mathcal{L}_{GT} + \eta $ to each image sample, where $\eta$ is a hyper-parameter. These values are used for the selection of triplet pseudo-labels.
To obtain pseudo-labels, we first run the standard HOI detector on the training set. For each detected result, we create prompt and compute the reconstruction loss in the same manner. Those with loss less than $\mathcal{L}_{GT} + \eta $ will be treated as pseudo-labels and included for subsequent fine-tuning of HOI detector.

\noindent\textbf{Image Generation.}
%In long-tail class distribution~\citep{logn,loga,tde,seesawloss,eqlv1,eqlv2}, a class whose frequency is below a certain threshold is considered rare, and the rare classes are usually pre-defined in the evaluation metrics.
The HOI datasets~\citep{hicodet} exhibit an extreme long-tail relation class distribution. As shown in Figure~\ref{fig:motivation_hico_longtail}, some classes have over 15,000 images, while many rare classes have fewer than 5 images.
% To select the best model for image generation, 
We conduct ablations on Textual Inversion~\citep{textual_inversion} and DreamBooth~\citep{dreambooth}, and find that DreamBooth yielded better results. Therefore, we train a DreamBooth model for each rare category and generated similar images. We add generated images to classes with fewer than 10 images to ensure there were at least 10 images for each class.
These newly generated images are not labeled, so we use the same label generation method to create labels. The textual annotations of images in each rare category are the same, so $\mathcal{L}_{GT}$ can be calculated directly using the triplet annotations of existing images as textual prompt.

\section{Experiments}
\label{sec:experiments}
We conduct experiments on HICO-DET~\citep{hicodet} and V-COCO~\citep{vcoco}.
In this section, we describe the datasets and evaluation settings, implementation details, ablation studies and comparisons to the state-of-the-art methods.

\subsection{Experimental Setting}
\label{sec:evaluation_settings}
\noindent\textbf{Datasets.} The models are evaluated on two public datasets: HICO-DET~\citep{hicodet} and V-COCO~\citep{vcoco}. HICO-DET has 47,776 images, and is split as 38,118 for training and 9,658 for testing. It contains 117 relation classes and 80 object categories. The relation and object classes can form 600 triplets, \emph{i.e.}, HOI categories. According to the frequency, these 600 HOI categories can be divided into 3 groups: Full (all HOI categories), Rare (138 HOI categories with fewer than 10 instances), and Non-Rare (462 categories with no fewer than 10 instances).
V-COCO is a subset of COCO, so it has the same 80 object classes as COCO. It contains 10,396 images with 5,400 images as the training split and 4964 images as the testing split. It has 29 relation classes, and among them, there are 4 body motions without any interaction with objects. Its quantity of the HOI triplets is 263.

\noindent\textbf{Zero-Shot Construction.} For zero-shot HOI detection, we follow the setting of previous work~\citep{GEN_VLKT}: Unseen Combination(UC), Unseen Object (UO), Rare First Unseen Combination (RF-UC), Non-rare First Unseen Combination (NF-UC), and Unseen Verb (UV).  
Specifically, the UC setting indicates the training data contains all categories of object and verb but misses some HOI triplet categories. We evaluate on the 120 unseen, 480 seen, and 600 full categories for the UC setting. 
The UO setting means the objects in the unseen triplets also do not appear in the training data. We use the unseen HOIs with 12 objects unseen among the total 80 objects and form 100 unseen and 500 seen HOIs for the UO setting. 
For UV, we randomly select 20 verbs from all total 117 verbs to form 84 unseen and 516 seen HOIs during training. 
Under the RF-UC setting, the tail HOI categories are selected as unseen classes, while the NF-UC uses head HOI categories as unseen classes. For RF-UC and NF-UC, we select 120 HOI categories as unseen classes.

\noindent\textbf{Evaluation Metric.} We use the same settings as \citep{qpic_hoi} and thus use the mean Average Precision (mAP) to measure our model. A detection result is considered as a true positive if the predicted human and object bounding box have an IoU higher than 0.5 with the corresponding ground-truth bounding boxes, and the predicted relation class is matched.
In HICO-DET, the object class is additionally used for evaluation, \emph{i.e.}, the object class of a prediction should match that of the ground-truth triplet.
We evaluate the models in two different settings: the default setting and the known-object setting. In the default setting, APs are calculated based on all the test images, while in the known-object setting, each AP is computed only based on images that contain the object category corresponding to each AP.
In V-COCO, as some HOIs are defined with no object labels, we evaluate the performance in two different scenarios following the official evaluation scheme of V-COCO. In scenario 1 (S1), the detectors report cases without any object. In scenario 2 (S2), the object predictions in these cases are ignored.

\noindent\textbf{Implementation Details.}
To verify the effectiveness of our CycleHOI training framework, we conduct experiments with various HOI detectors. To ensure the fairness of the experiments, we do not alter the configuration of these HOI detectors and use their official code. The network structure and hyper-parameters of these detectors remain unchanged.
The loss weights $\lambda_{Det}$, $\lambda_{Cycle}$ and $\lambda_{Dis}$ are set to $1$, $0.2$ and $10$, respectively.
For the standard Stable Diffusion~\citep{stable_diffusion}, as well as its applications Textual Inversion~\citep{textual_inversion} and DreamBooth~\citep{dreambooth}, we use the pre-training weights from its v1.5 version. We conduct all the experiments with a batch size of 16 on 8 NVIDIA
A100 GPUs with 80GB of memory.
We evaluate the performance of the proposed method on HICO-DET and V-COCO using the evaluation codes from the QPIC~\citep{qpic_hoi}.

\subsection{Results on the Regular HOI Detection}
\label{sec:regular_det}
The results on the datasets of HICO-DET and V-COCO are presented in Table~\ref{tab:sota_hico} and Table~\ref{tab:sota_coco}. To validate the effectiveness of our CycleHOI, we conduct experiments on a variety of HOI detectors using different backbones, including ResNet-50, ResNet-101, and Swin-L. For HICO-DET, our CycleHOI provides a stable boost on various DETR-based HOI detectors. The performance improvement of our CycleHOI training framework is not so evident for large backbone (e.g., Swin-L) as smaller backbone (e.g., ResNet-50). This is due to the fact that a larger backbone will have a stronger modeling and characterization capability, and its performance is much higher than smaller ones. In addition, it can also be seen from the experimental results that there is a large enhancement for the rare categories as we specifically mitigate the long-tail problem of the dataset. For V-COCO, our CycleHOI also achieves a similar performance improvement, which confirms the generalization ability of our method. Finally, we find that our CycleHOI with PViC detector obtains the state-of-the-art performance on two datasets.

\hicosotatable

\cocosotatable

\subsection{Results on the Zero-Shot HOI Detection}
\label{sec:zero_shot_det}
We use the pre-trained diffusion model to improve the performance of the HOI detector, so its zero-shot capability is also worth exploring. We use GEN-VLKT~\citep{GEN_VLKT} as a baseline to verify the effectiveness of CycleHOI on zero-shot HOI detection. The experimental results are shown in Table~\ref{tab:zero_shot}. It can be seen that after adding the proposed methods, there can be a great improvement in UC, UO and UV, and it can outperform the state-of-the-art methods in some of the metrics. This demonstrates that the diffusion model can substantially improve the zero-shot capability of HOI detector. We also for some settings, our method is inferior to previous HOICLIP~\citep{hoiclip}, mainly due to their specific zero-shot design in the detector pipeline, which is out the scope of our paper. In the future, we could consider combining our CycleHOI with HOICLIP.

\zeroshottable

\subsection{Ablation Studies}
\label{sec:ablation}
In this section, we conduct in-depth ablations to explore the optimal experimental setting and analyze the effectiveness of the our CycleHOI.

\noindent\textbf{Effectiveness of Proposed Techniques.} 
Table~\ref{tab:module_benefit} gives a detailed ablations on the proposed modules in our method. Specifically, we investigate the effectiveness of cycle consistency (CC), knowledge distillation (KD), and dataset enhancement (DE) in a step-by-step manner. Overall, these techniques are complimentary to each other and each contributes to a better performance. The cycle consistency loss obtains the best improvement among three techniques (30.44 (CC) vs. 30.01 (KD) vs. 30.09 (DE)). When combining all these tehcniques, our CycleHOI can boost the final performance to 32.23 mAP.

\noindent\textbf{U-Net Feature Map Setting.} 
The U-Net module of Stable Diffusion has a total of 3 stages of up-sampling, starting from the middle $8 \times 8$ sized feature map gradually performing 2$\times$ up-sampling and finally reaching $64 \times 64$ sized feature maps, denoted as \textbf{Stage0-Stage3} in that order. Therefore we try to explore specifically which stage of the feature map to use for distillation works best. The experimental results are shown in Table~\ref{tab:ablation:distill_size}, where \textbf{All} indicates that the feature maps of these 4 stages are fused according to FPN~\citep{fpn}. From the results, distillation using the last stage of the feature map is the most effective, boosting 1.05 mAP.

\noindent\textbf{Time Step Setting.} 
In Stable Diffusion, time step controls the granularity of image generation. When time step is small, the granularity of image is coarser and easier to be controlled by the text. As the time step becomes larger, it pays more attention on the details. Therefore, it is important to determine the appropriate time step to generate the feature map of distillation, and the experimental results are shown in Table~\ref{tab:ablation:distill_time}. The best result is obtained when time step is 1.

\ablationtable

\noindent\textbf{Threshold Setting.} 
Filtering pseudo-labels according to a threshold $\eta$ is shown in section~\ref{sec:dataset}. Table~\ref{tab:ablation:label_thresh} gives the performance improvement of filtering pseudo-labels at different thresholds $\eta$. The performance increases first and then decreases with the increasing of the threshold, and reaches the maximum when the threshold is set to 1.

\noindent\textbf{Cycle Consistency Loss Setting.} 
First, we studied which loss function is more effective for calculating cycle consistency loss, conducting experiments using L1 loss, L2 loss, and perceptual loss(\textbf{PL}), with results shown in Table~\ref{tab:ablation:ti_loss}. As can be seen from the table, fine-grained loss function like L1 and L2 loss perform better. This is because they can optimize the replaced embeddings more effectively, allowing for the generation of more accurate bounding boxes and classes. Besides, we have three ways of calculating the loss when adding cycle consistency constraints, denoted as \textbf{M1-M3}, as shown in Table~\ref{tab:ablation:ti_method}. \textbf{M1}: we compose all positive embeddings into a one-sentence embedding to be fed into the generator and supervise it with the full image. This is the implementation that works best and our default approach, as illustrated in Figure~\ref{fig:framework}. The loss function is: $\mathcal{L}_{Cycle} = \|g([L; V_1; V_2; \cdots; V_M]) - \boldsymbol{I}\|_2$, where $g$ denotes the generator and $L$ denotes the word embeddings of ``\textit{A photo of}''.
\textbf{M2}: we compute the loss once for each positive embedding and supervise it with the full image. Then the loss is: $\mathcal{L}_{Cycle} = \|g([L; V_1]) - \boldsymbol{I}\|_2 + \cdots + \|g([L; V_M]) - \boldsymbol{I}\|_2$. \textbf{M3}: we compute the loss once for each positive embedding and supervise it separately with the corresponding image region indicated by the ground-truth that matches it. Then the loss is:
$\mathcal{L}_{Cycle} = \|g([L; V_1]) - \boldsymbol{I}_1\|_2 + \cdots + \|g([L; V_M]) - \boldsymbol{I}_M\|_2$, 
where $\boldsymbol{I}_1$-$\boldsymbol{I}_M$ denote the corresponding image regions, and we take the union box of the human box and the object box. From the results, we see that the \textbf{M1} loss form achieves the best performance.

\noindent\textbf{Generation Model Setting.} 
In section~\ref{sec:dataset}, similar images need to be generated for rare categories. There are several methods for generating images of personalized concepts, including Textual Inversion~\citep{textual_inversion} and DreamBooth~\citep{dreambooth}. To explore which of these two methods can provide a better understanding of the concept of rare categories, we conduct a comparison experiment and the results are displayed in Table~\ref{tab:ablation:long_tail}. From the results, we see that the DreamBooth achieves the better performance.

\section{Conclusion}
In this paper, we have presented an enhanced training framework, coined as CycleHOI, to improve the performance of learned HOI detector by bridging the powerful pre-trained text-to-image diffusion model with the popular DETR detection pipeline. We introduce a novel cycle consistency loss over the processes of instance decoding and instance inversion to encourage the detected HOI instances to be able to reconstruct the original image. In addition, we design an one-step denoising strategy to transfer diffusion model representation to the DETR encoder via knowledge distillation. From a more practical view, we also augment the training set with diffusion models from both aspects of label correction and data generation. The experiment results demonstrate the effectiveness of our CycleHOI on improving HOI detector without introducing any extra inference cost.

\bibliographystyle{main}
\bibliography{main}

% \section{Appendix}

\end{document}